\documentclass[3p]{elsarticle}

\usepackage{lineno,hyperref}

\usepackage{amssymb}
\usepackage{amssymb}
\usepackage{epsfig}
\usepackage{subfigure}
\usepackage{graphics}
\usepackage{graphicx}
\usepackage{amsmath}
\usepackage{diagbox}
\usepackage[table]{xcolor}
\usepackage{multirow}
\usepackage{array}
\usepackage{indentfirst}
\usepackage{caption}
\usepackage{makecell}
\usepackage{rotating}

\usepackage{natbib}
\usepackage{verbatim}
\usepackage{booktabs}

\usepackage{algorithm}
\usepackage{algpseudocode}

\journal{Information Sciences}

\begin{document}

\begin{frontmatter}

\title{Distributional Discrepancy: A Metric for Unconditional Text Generation}
\author[mymainaddress]{Ping Cai\fnref{equalcontribution}}
\ead{caiping@my.swjtu.edu.cn}

\author[mysecondaryaddress]{Xingyuan Chen\fnref{equalcontribution}}
\fntext[equalcontribution]{The authors contributed equally to this work.}
\ead{chenxy@nlp.nju.edu.cn}

\author[mysecondaryaddress]{Peng Jin\corref{mycorrespondingauthor}}
\ead{jandp@pku.edu.cn}

\author[mymainaddress]{Hongjun Wang\corref{mycorrespondingauthor}}
\cortext[mycorrespondingauthor]{Corresponding author}
\ead{wanghongjun@swjtu.edu.cn}

\author[mymainaddress]{Tianrui Li}
\ead{trli@swjtu.edu.cn}

\address[mymainaddress]{School of Information Science and Technology, Southwest Jiaotong University}
\address[mysecondaryaddress]{School of Computer Science, Leshan Normal University}

\begin{abstract}
The purpose of unconditional text generation is to train a model with real sentences, then generate novel sentences of the same \textit{quality} and \textit{diversity} as the training data. However, when different metrics are used for comparing the methods of unconditional text generation, contradictory conclusions are drawn. The difficulty is that both the diversity and quality of the sample should be considered simultaneously when the models are evaluated. To solve this problem, a novel metric of distributional discrepancy (DD) is designed to evaluate generators based on the discrepancy between the generated and real training sentences. However, it cannot compute the DD directly because the distribution of real sentences is unavailable. Thus, we propose a method for estimating the DD by training a neural-network-based text classifier. For comparison, three existing metrics, bi-lingual evaluation understudy (BLEU) versus self-BLEU, language model score versus reverse language model score, and Fr\'{e}chet embedding distance, along with the proposed DD, are used to evaluate two popular generative models of long short-term memory and generative pretrained transformer 2 on both syntactic and real data. Experimental results show that DD is significantly better than the three existing metrics for ranking these generative models.
\end{abstract}


\end{frontmatter}


\section{Introduction}
Unconditional text generation, as the cornerstone of conditional text generation such as dialogue generation \citep{Li2017Adversarial}, machine translation \citep{Wu2016Google}, and image caption \citep{xu2015show}, has been studied extensively \citep{Yu2016SeqGAN,Caccia2019falling,dAutu2019Scratch}. In this task, a model is usually fed with many real independent sentences and then required to generate numerous novel independent sentences that should have the same \textit{quality} and \textit{diversity} as the training data. Neural language models (LMs), such as recurrent neural network (RNN)-based \citep{hochreiter1997long} and transformer-based \citep{Vaswani2017Attention} models, and their variations, which are trained using maximum likelihood estimation (MLE), are used to achieve this task. Owing to exposure bias in the reference stage \citep{Bengio2015Scheduled}, a generative adversarial network (GAN) \citep{Goodfellow2014Generative} is introduced to solve this problem, and a flooding of variants occur \citep{Lin2017Adversarial,Fedus2018MaskGAN,Nie2019ICLR}. However, when these considered language GANs are evaluated precisely in terms of both quality and diversity of the sample, the results show that they are easily beaten by simply adjusting the softmax temperature of the long short-term memory (LSTM) \citep{Shi2018Toward,Semeniuta2018On,Caccia2019falling}. Therefore, a good evaluation metric is imperative for the research community.

It seems that a human evaluation is the best choice. Regardless of the expense and non-repeatability, humans can only evaluate the quality of a single sentence, and cannot accurately determine the diversity when hundreds of thousands of sentences are presented\citep{Hashimoto2019HUSE}.

A few automatic methods have been proposed. In \cite{Yu2016SeqGAN}, the authors adapted bilingual evaluation understudy (BLEU) \citep{Papineni2002BLEU} for measuring the quality, although it can only capture the local consistency \citep{dAutu2019Scratch}. As its counterpart, self-BLEU was introduced for measuring the diversity \citep{Zhu2018Texygen}. This two-dimensional metric fails when one model is better than the other in quality but worse in diversity, or vice versa. A similar situation is to apply another paired metric, namely, LM score versus reverse LM score \citep{DBLP:journals/corr/abs-1804-07972}. To accurately measure the models, in \cite{Caccia2019falling}, the authors draw several dots in the two-dimensional space by adjusting the softmax temperature and then linking them using a line. This method requires a considerable number of computations, and unfortunately, its ability to discriminate is poor. The Fr\'{e}chet embedding distance (FED), which was first used in image generation, was proposed as a single metric to evaluate unconditional text generation models \cite{dAutu2019Scratch}. It cannot yet accurately evaluate several generators with a temperature of 1.0. When adjusting the temperature, some models are better than others when temperatures are less than 1.0; once the temperature is higher than 1.0, these models become worse than the others. Moreover, the BERTScore \citep{Zhang2020BertScore} improved the quality measure by embedding each token into a low-dimensional space but still neglected the sample diversity. In \cite{Hashimoto2019HUSE}, a framework for unifying human and statistical frames was developed; however, it requires at least 10 crowd-workers, thereby making it an expensive metric that is difficult to reproduce.



There is a difference between real and generated texts. In \cite{Zellers2019fakenews}, a transformer-based neural LM was trained with 120 GB of human-written news. Although the accuracy of human prediction of real or fake news is approximately 70\%, the accuracy of a well-trained classifier is more than 95\%. This means a text classifier detects discrepancies extremely well. Thus, we propose a novel metric, namely, distributional discrepancy (DD), to measure the discrepancies between these two sets of texts. When the DD score of a generator is smaller, the distribution of sentences generated is closer to the distribution of real sentences. This implies that this generator is better.

It is impossible to accurately compute the DD score because the distribution of real texts cannot be directly obtained. We propose a learning method to estimate the DD as inspired by \citep{Lopez-Paz2017Tests}, and \citep{Lopez-Paz2017Tests} assess whether two samples are drawn from the same distribution. We use a text classifier trained with both generated and real texts to detect the discrepancies between them. A discrepancy is not used to discriminate whether these two sets belong to the same distribution, but rather to measure the DD between them.

%

%



Our contributions are as follows:

\begin{itemize}
\item The discrepancy between real and generated texts is used to evaluate the performance of a generative model, and the DD, as a single metric, can simultaneously measure the quality and diversity of the sample.
%

\item A neural network-based text classifier can be trained to estimate the DD, and the discrepancy can be computed according to the performance of this classifier.

%

\item Two popular neural LMs, LSTM and generative pretrained transformer 2 (GPT-2), are applied to synthetic and real data, and the experimental results show that the rank by the DD corresponds to the gold-standard order, and the three existing metrics fail to rank the generators.

\end{itemize}

The remainder of this paper is organised as follows. Section 2 presents related works. The novel single-metric DD is defined in Section 3. To estimate the DD, the implementation procedure as a learning method is introduced in Section 4. In Sections 5 and 6, we detail the evaluations of 10 unconditional text generators on synthetic and real data, respectively. Finally, conclusions are provided in Section 7. All the codes and datasets are available at \url{https://github.com/anonymous1100/Distributional-Discrepancy}.

\section{Related Work}
Several widely used metrics for unconditional text generation are introduced in this section. The methods for applying a text classifier for this task are described in further detail.

\subsection{Existing Metrics}
The quality of the generated sentences is considered the most important aspect. BLEU \citep{Papineni2002BLEU}, which evaluates the quality of a translated sentence given a source sentence, has been used as a metric by researchers\citep{Yu2016SeqGAN}. It is undoubtedly appropriate to evaluate machine translation models because the translated sentences only need to be compared with a few reference translations provided by experts. However, there are usually hundreds of thousands of real sentences as references in unconditional text generation. Another drawback is that BELU only measures the local consistency. Although the LM score can capture global semantics, it biases those models that generate highly likely sentences \citep{Semeniuta2018On}.

The diversity of the generated sentences is as important as the quality when evaluating an unconditional generator. In \cite{Zhu2018Texygen}, the authors proposed self-BLEU, which calculates the BLEU among those generated sentences. The reverse LM score was first used by \cite{Zhao2018Autoencoders}. Owing to the modelling imperfection and the training data bias, this is not a good proxy for a model's diversity \cite{Semeniuta2018On}.

A natural idea is using BLEU versus self-BLEU as a paired metric to evaluate an unconditional text generation in terms of both quality and diversity simultaneously \citep{DBLP:journals/corr/abs-1804-07972}. However, in \cite{dAutu2019Scratch}, a simple 5-gram LM with Kneser-Ney smoothing \citep{Kneser1995smooth} was thought to perform nearly perfectly, although in fact it generates extremely poor quality sentences. Considering such limits, in \cite{DBLP:journals/corr/abs-1804-07972}, LM score versus reverse LM score were proposed to evaluate the quality and diversity, respectively. Owing to a shortage of paired metrics, we must obtain several values by adjusting the softmax temperature when one model is better than another in quality but worse in diversity, or vice versa \citep{Caccia2019falling}.

Owing to the inconvenience of a paired metric, in \cite{dAutu2019Scratch} the FED is proposed as a single metric that originates from the image generation \citep{Heusel2017Nash}. It claims to capture the global consistency and is faster than BLEU. However, a temperature adjustment is unavoidable because it fails to discriminate generators under the condition of a temperature of 1.0.



\cite{Hashimoto2019HUSE} proposed the use of Human and Statistical Evaluation as a metric combining a human evaluation and a statistical approach to approximate the probability under a real text distribution. They then train a simple $k$-nearest neighbour classifier and apply the leave-one-out error of this classifier twice for determining the discrepancy between real and generated text. The limitation here is the requirement of the crowd-workers. Similar to this approach, we train a classifier but without any human labour.

The latest metric is BERTScore \citep{Zhang2020BertScore}. By computing the similarity of the tokens in a contextual embedding space, this approach can measure the sample quality better than BLEU. Regretfully, however, the sample diversity is neglected in this case. Of course, there are many other measures such as K-L divergence, but all require knowing the explicit distributions, which is clearly unrealistic because we cannot know the distribution of the real texts.

\subsection{Using Text Classifier to Detect Discrepancy}
GANs \citep{Goodfellow2014Generative} have improved the training of a neural LM by fine-tuning it \citep{Yu2016SeqGAN,Fedus2018MaskGAN,Nie2019ICLR}. With language GANs, a discriminator operates as a classifier to detect the discrepancy between real sentences and sentences generated by the most recent generator. This detected signal is fed to the generator. To avoid the local optima, the discriminator is usually applied for several epochs during the adversarial learning. However, with the goal of evaluating the generators, we try to train a convenience classifier with many epochs to obtain an approximation of the optimal classifier. Clearly, both aims and methods are different from our own.

\cite{Lopez-Paz2017Tests} used a classifier to detect the discrepancy between two samples to assess whether they are drawn from the same distribution. They construct a dataset consisting of equal examples from these two samples. The examples from one sample are labelled as positive and the other as negative. A binary neural network classifier is trained using these examples. If the classification accuracy on the held-out data approximates to 0.5, these two samples are classified as complying with the same distribution. Otherwise their distributions are different from each other. Unlike \cite{Lopez-Paz2017Tests}, we use a text classifier, which is trained with generated and real texts, to detect this discrepancy. This discrepancy is not used to discriminate whether these two sets belong to the same distribution, but to measure their distributional discrepancy.

\section{Distributional Discrepancy}
Given a set of real sentences $\mathcal{T}_r$, $x \in \mathcal{T}_r$, $x=[\mathrm{x}_1,...,\mathrm{x}_L]$ is a sentence of length $L$, and $\mathrm{x}_i$ is the i-th word, $x \sim p_r(x)$. An unconditional text generator $G_{\theta}$ is trained using $\mathcal{T}_r$, and then generates a set of sentences $\mathcal{T_{\theta}}$. As a sentence, $x$ is generated by $G_{\theta}$,  $x \sim p_\theta(x)$. The closer $p_\theta(x)$ is to $p_r(x)$, the better $G_{\theta}$ becomes. Therefore, the discrepancy between $p_\theta(x)$ and $p_r(x)$ can be used as a metric to evaluate the generative model.



We propose the DD to measure this discrepancy. This metric is defined as follow:

\begin{equation}\label{equDs}
d_d = \frac{1}{2}\int{\big|p_r(x) - p_\theta(x)\big|}dx
\end{equation}

\noindent where $x \in \Omega$ ($\Omega$ is the space of all possible samples).

Clearly, the range of this function is $0\sim1$. 

Unfortunately, the mathematical form of $p_r(x)$ cannot be obtained. To detect this discrepancy as precisely as possible, we propose a learning method to estimate this function.

\subsection{Learning Method for Obtaining Distributional Discrepancy}
To transform the computation of the DD into a learning method, equation \ref{equDs} is inferred as follows:

\begin{equation}
\begin{split}
\label{equDsCal}
d_d &= \frac{1}{2}\int{\big|p_r(x) - p_\theta(x)\big|dx} \\
    &= \frac{1}{2}\bigg[\int_{p_r(x)\geq p_\theta(x)}{\big(p_r(x) - p_\theta(x)\big)dx} + \int_{p_r(x)<p_\theta(x)}{\big(p_\theta(x) - p_r(x)\big)dx} \bigg]\\
    &= \frac{1}{2}\bigg[\int_{p_r(x)\geq p_\theta(x)}{p_r(x)dx} + \int_{p_r(x)<p_\theta(x)}{p_\theta(x)dx} - \int_{p_r(x)\geq p_\theta(x)}{p_\theta(x)dx} - \int_{p_r(x)< p_\theta(x)}{p_r(x)dx} \bigg]\\
    &= \frac{1}{2}\bigg[\mathbb{E}_{\substack{x\sim{p_r(x)}\\ {p_r(x)\geq p_\theta(x)}}}\big(1\big)  +
    	\mathbb{E}_{\substack{x\sim{p_\theta(x)}\\ {p_r(x)<p_\theta(x)}}}\big(1\big) -
    	\mathbb{E}_{\substack{x\sim{p_\theta(x)}\\ {p_r(x)\geq p_\theta(x)}}}\big(1\big) -
    	\mathbb{E}_{\substack{x\sim{p_r(x)}\\ {p_r(x)<p_\theta(x)}}}\big(1\big) \bigg]\\
    &= \frac{1}{2}\bigg[\mathbb{E}_{\substack{x\sim{p_r(x)}\\ {z\geq 0.5}}}\big(1\big)  +
    	\mathbb{E}_{\substack{x\sim{p_\theta(x)}\\ {z<0.5}}}\big(1\big) -
    	\mathbb{E}_{\substack{x\sim{p_\theta(x)}\\ {z\geq 0.5}}}\big(1\big) -
    	\mathbb{E}_{\substack{x\sim{p_r(x)}\\ {z<0.5}}}\big(1\big) \bigg]\\
\end{split}
\end{equation}

where $z = \frac{p_{r}(x)}{p_{r}(x) + p_\theta(x)}$.


The computation of DD is transformed to resolve $z$, which can be obtained using the learning method. Given $\theta$, according to \citep{Goodfellow2014Generative}, to detect the discrepancy between $p_\theta(x)$ and $p_{r}(x)$, $D_\phi$ is defined and optimised as follows:

\begin{equation}\label{equOptimD}
\mathop{max}\limits_{D_\phi}V(D_\phi, G_\theta) = \mathop{max}\limits_{D_\phi} \mathbb{E}_{x\sim{p_{r}}}\bigg[{logD_\phi(x)}\bigg]+\mathbb{E}_{x_\sim{p_\theta}}\bigg[{log\big(1 - D_\phi(x)\big)}\bigg]
\end{equation}

Assuming that $D_{\phi}^{*}(x)$ is the optimal solution, we have the following:

\begin{equation}\label{equBestD}
D_{\phi}^{*}(x) = \frac{p_{r}(x)}{p_{r}(x) + p_\theta(x)}
\end{equation}

Thus, $z = D_{\phi}^{*}(x)$, and

\begin{equation}\label{Equ0.5}
    \begin{cases}
    D_{\phi}^{*}(x) \geq 0.5,   \qquad iif \quad p_r(x) \geq p_\theta(x)\\[2ex]
    D_{\phi}^{*}(x) < 0.5,   \qquad iif \quad p_r(x) < p_\theta(x)
    \end{cases}
\end{equation}

\subsection{An Estimation Function of Distributional Discrepancy}


In this subsection, an estimation function of DD is illustrated. According to equation \ref{Equ0.5}, the integration of the density function can be transformed into a statistical function. Substituting $z$ in equation \ref{equDsCal} with equation \ref{equBestD}, we have the following:

\begin{equation}\label{equ8}
    	d_d= \frac{1}{2}\bigg[
    	\mathbb{E}_{\substack{x\sim{p_r(x)}\\ D_{\phi}^{*}(x)>0.5}}\big(1\big) -
    	\mathbb{E}_{\substack{x\sim{p_r(x)}\\ D_{\phi}^{*}(x)\leq0.5}}\big(1\big) +
    	\mathbb{E}_{\substack{x\sim{p_\theta(x)}\\ D_{\phi}^{*}(x)\leq0.5}}\big(1\big) -
    	\mathbb{E}_{\substack{x\sim{p_\theta(x)}\\ D_{\phi}^{*}(x)>0.5}}\big(1\big)
    	\bigg]
\end{equation}

Assuming that the classification accuracy of $D_{\phi}^{*}$ is $a$, the classification error is $b = 1-a$. According to equation \ref{equ8}, $d_{d} = a-b = 2*a-1$. When $p_r(x) \equiv p_\theta(x)$, $d_d = 0$.


In fact, it is critical to obtain the optimal $D_{\phi}^{*}$ and approximate its accuracy to 1. Fortunately, neural text classifiers are substantially powerful \citep{Kim2014CNN,Lai2015RCNN}. Thus, a learning method to approximate $D_{\phi}^{*}$ is practicable by training this classifier. In the next section, DD can be estimated using a text classifier.


\begin{figure*}[htbp]
\centering

\includegraphics[scale=0.5]{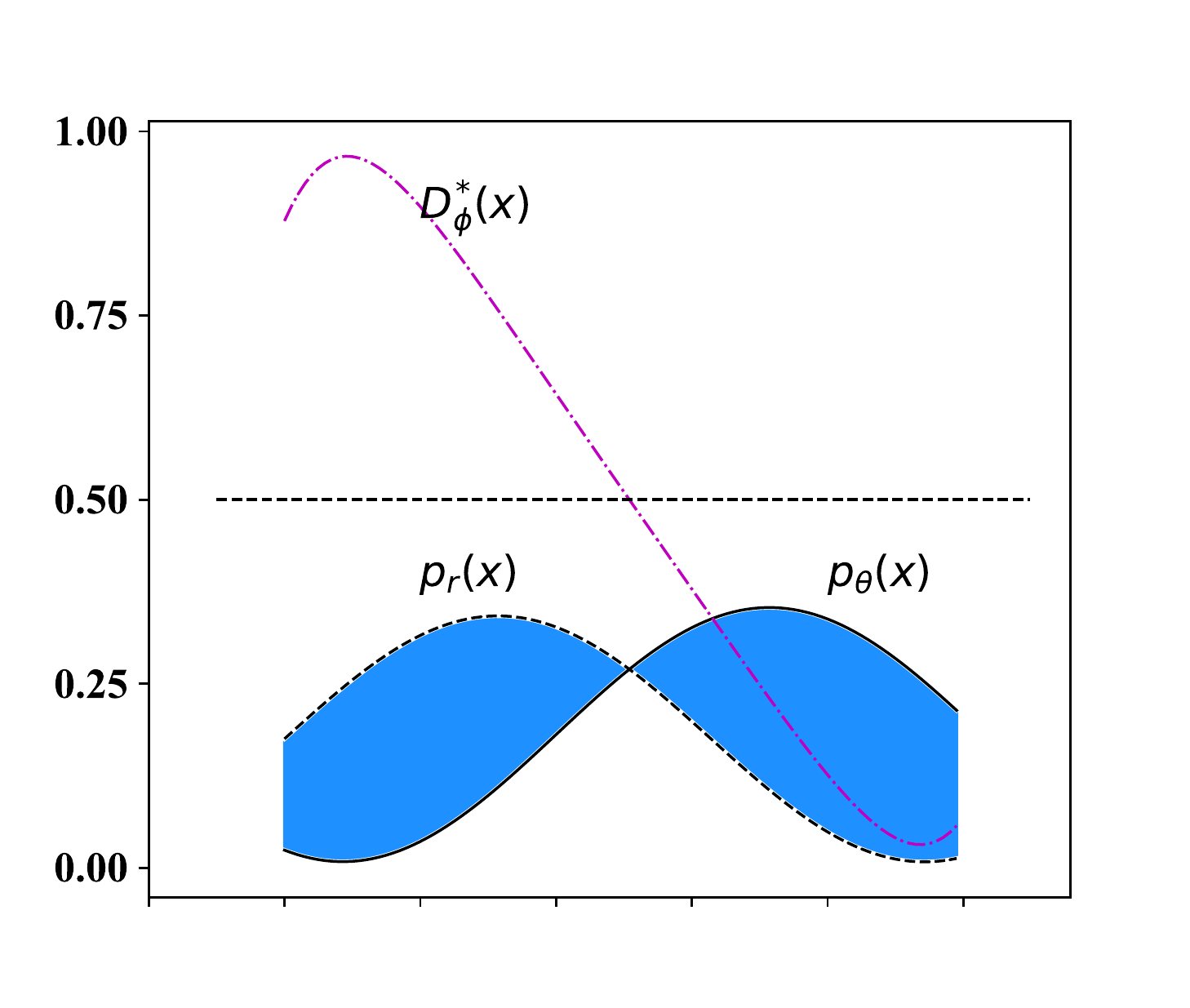}

\caption{Illustration of distributional discrepancy. Half of the shaded area equals the result of equation \ref{equ8}. Larger values indicate a greater discrepancy.}
\label{figDiscrepancy}
\end{figure*}

\section{Estimating the Distributional Discrepancy Using a Learning Method}
In equation \ref{equ8}, the optimal classifier function $D_{\phi}^{*}$ can only be statistically estimated by an approximated function. We adapt a widely used convolutional neural network (CNN)-based text classifier \citep{Kim2014CNN} and denote it as $D_\phi$\footnote{Recently, it was claimed that the transformer-based classifier achieves the highest accuracy. We will try to apply it in a future study. Our experimental results show that a CNN works well.}. Here, $D_\phi$ is trained using the samples from real and generated sentences, according to equation \ref{equOptimD}. When it convergences, we obtain $\hat{D}_\phi$. In addition, $\hat{D}_\phi$ is an approximation of $D_{\phi}^{*}$, and thus the estimation of $d_d$ can be computed. The degree of approximation is mainly determined by three factors: the structure and number of parameters of $D_{\phi}$, the volume of the training data, and the settings of the hyper-parameters. The procedure for estimating the discrepancy is described as follows.

\begin{flushleft}
\qquad Step 1: Design a discriminator $D_\phi$. This is usually a neural network such as a CNN.

\qquad Step 2: All real sentences in $\mathcal{T}_r$ are labelled as positive, and all generated sentences in $\mathcal{T_\theta}$ are labelled as negative. To avoid imbalanced learning \citep{He2009Imbalanced}, we let $G_\theta$ generate the same number of sentences as $\mathcal{T}_r$. A training set, which is denoted as $\mathcal{T}_{train}$, is composed of the same number of sentences selected randomly from these two sets. For example, 80\% of the sentences are selected. Further, 10\% of the sentences consist of a validation set $\mathcal{T}_{dev}$, and test set $\mathcal{T}_{test}$ is composed of the remaining 10\% of the samples.

\qquad Step3: Here, $D_\phi$ is trained with $\mathcal{T}_{train}$ and optimised according to equation \ref{equOptimD}. It should be noted that the training strategy is significantly different from that of a GAN. We do not stop training $D_\phi$ until its classification accuracy on $\mathcal{T}_{dev}$ convergences. This usually requires $50\sim100$ epochs\footnote{The number of epochs is mainly determined by the learning ratio.}. A GAN applies numerous adversarial rounds and each round trains several epochs \citep{Yu2016SeqGAN} to avoid becoming trapped in the local optima. We denote this converged $D_\phi$ as $\hat{D}_\phi$, which is an approximation of $D_{\phi}^{*}$.

\qquad Step4: According to equation \ref{equ8}, we can compute the discrepancy between real and generated sentences set by the prediction of $\hat{D}_\phi$ on $\mathcal{T}_{test}$. This is denoted as $\hat{d}_d$ for an estimation of the real DD.
\end{flushleft}


This procedure is a learning method because $\hat{D}_\phi$ as the estimation of $D_{\phi}^{*}$ is learned using a neural network.  A well-trained $\hat{D}_\phi$ can be a meaningful approximation of $D_{\phi}^{*}$. Therefore, $\hat{d}_d$ is obtained as the meaningful approximation of $d_d$ through $\hat{D}_\phi$. Although $d_d \ne \hat{d}_d$, their tendency to change is the same. If $\hat{d}_d$ of a generator is smaller, this generator is better.


\section{Experimental Setup}
To verify the DD metric, we need several unconditional text generation models and must obtain their gold-standard in advance. These generators generate sentences and are ranked according to our novel metric. Meanwhile, they are also ordered by three baseline metrics. A correlation coefficient with the gold-standard order will be computed for a comparison of all metrics.


\subsection{Dataset}
A benchmark dataset, EMNLP2017 WMT news\footnote{http://www.statmt.org/wmt17/}, for unconditional text generation is used as the real corpus. In this corpus, the average length of a sentence is approximately 20 words. There are a total of 5255 word types, and the longest sentence consists of 51 words. All training data, approximately 280,000 sentences, are used and there are 10,000 sentences in the test data. It should be noted that we separate the last 10,000 sentences from the training data as the validation set.
Considering the impossibility of obtaining the distribution of real texts, we experiment further with synthetic data. A state-of-the-art neural LM, GPT-2, is trained using EMNLP2017 WMT News \footnote{To distinguish this GPT-2 model from the latter models, it is called the original GPT-2 generator.}. Then, the sentences generated by this model as the \textit{real} data, consist of 320,000 sentences in total. Among them, we divide the first 300,000 sentences as training data, middle 10,000 sentences as the validation set, and last 10,000 sentences for testing.

\subsection{Generator and Classifier}
Two widely used neural LMs: LSTM (RNN-based architecture) and GPT-2 (transformer-based architecture), which are trained according to the MLE, are evaluated as the generative models. The left side of Table \ref{hyper} lists their hyper-parameters.

\begin{table}[t!]
\centering
\begin{tabular}{|c|c|c|c|}
\hline
{H.P. of Generator} & {Value} & {H.P. of Classifier} & {Value}\\
\hline
hidden size & 512 (768) & layer1 & (2, 100)\\
layer & 2 (12)  & layer2 & (3, 200)\\
drop\_out &0.5 (0.1) & drop\_out &0.5\\
learning rate & 1e-3 (2e-5) & learning rate & {1e-4}\\
{batch size} & {128} &   {batch size} & {512} \\
{GPT-2 head}  & {12} & -- & -- \\
{number of para.}  & {$\approx$30.1(268)M} & {number of para.} & {$\approx$10.5M} \\
\hline

\end{tabular}
\caption{Values of hyper-parameters. For two generative models, the values of GPT-2 are listed in parentheses when they are different from those of the LSTM. `H.P.' indicates a hyper-parameter and `number of para' denotes the total number of parameters. For each convolutional layer, (window size, kernel numbers) are listed.}
\label{hyper}
\end{table}

For the classifier, we use a CNN whose hyper-parameters are the same as those of the discriminator used in \citep{Yu2016SeqGAN}. The right side of table \ref{hyper} lists the hyper-parameters. The training data consist of positive samples used to train the generator, and the same number of negative samples that are generated by the generator. We do not stop training the classifier until a convergence of the classification accuracy is observed on the validation set. Finally, the one with the highest accuracy on the verification set is used for a prediction of the test set.

For each generator, regardless of the volume of the training data or model architecture, we always let it generate 320,000 sentences, which are labelled as negative samples. Among them, the first 300,000 sentences are used as the classifier's training set. The middle and last 10,000 sentences are used as the verification and test sets, respectively. It should be noted that all positive sentences that are combined into these three sets are generated by the original GPT-2 generator.

\subsection{Gold-standard Order}
According to the experience from vision tasks \citep{sun2017data}, given the model type and the settings of the hyper-parameters, the gold-standard order should be the same as the rank of the volume of the dataset, i.e., the more data that are applied, the better the performance.

For a real corpus, the training data are divided into 20$\%$, 40$\%$, 60$\%$, 80$\%$, and 100$\%$ sets from the first sentence to the last. Therefore, five datasets are obtained and the larger one always contains a smaller one.  Both LSTM and GPT-2 are trained using these five training datasets \footnote{In practice, we train 80 epochs and select the one that achieves the lowest PPL score on the validation set as the generator for comparison.}. Thus, five LSTM generators and five GPT-2 generators are created.

The drawback of using real data is that we cannot compare different architecture models. As described in the previous section, an original GPT-2 generates real sentences to train other generators for evaluation. We also divide the training data into five parts using the same partition as the real data. Unlike a real scenario, because $p_r(x)$ can be obtained from the original GPT-2, we directly compute the DD using equation \ref{equDsCal}. All 10 generators are ranked according to this DD score as the gold-standard, as shown in Table \ref{Tb:performanceBYequation}.

Given the same training data, GPT-2-based generators are always better than LSTM based generators because the number of parameters of GPT-2 is approximately seven times that of the latter, and its architecture is better than that of LSTM \citep{Radford19}.

\begin{table}[h]
  \caption{Evaluation of generators on syntactic datasets. The model with a digit as its footnote denotes a generator trained using the ratio to the entire \textit{real} dataset. Here, $dd$ is the DD score directly computed using equation \ref{equDsCal}.Accuracy is the classification accuracy and $\hat{dd}$ is the estimation of $dd$. For all cases, the lower the value is, the better the performance.}
   \label{Tb:performanceBYequation}
   \begin{center}
   \begin{tabular}{|c|ccc||c|ccc|}
   \hline
    Generator & $dd\downarrow$  & Accuracy & $\hat{dd\downarrow}$ & Generator & $dd\downarrow$  & Accuracy & $\hat{dd\downarrow}$ \\
   \hline
   $\text{LSTM}_{0.2}$ & 0.994 & 0.721 & 0.442 & $\text{GPT-2}_{0.4}$ & 0.973 & 0.644 & 0.287 \\
   \hline
    $\text{GPT-2}_{0.2}$ & 0.991 & 0.691 & 0.381 & $\text{LSTM}_{1.0}$ & 0.972 & 0.628 & 0.256 \\
   \hline
    $\text{LSTM}_{0.4}$ & 0.986 & 0.679 & 0.359 & $\text{GPT-2}_{0.6}$ & 0.959 & 0.621 & 0.241\\
   \hline
    $\text{LSTM}_{0.6}$ & 0.980 & 0.657 & 0.314 & $\text{GPT-2}_{0.8}$ & 0.948 & 0.609 & 0.218\\
   \hline
    $\text{LSTM}_{0.8}$ & 0.976 & 0.644 & 0.289 & $\text{GPT-2}_{1.0}$ & 0.935 & 0.596 & 0.192\\
   \hline

 \end{tabular}
 \end{center}
\end{table}

\subsection{Baseline Metrics}
Three single metrics, namely, BLEU, LM score, and FED, are compared as the baseline metrics. For BLEU, we use a 5-gram as the implementation. Further comparison by adjusting the softmax temperature is described in the next section. The perplexity is not adapted because we must compare the quality and diversity of the generated sentences against the real values, rather than observing the performance on real sentences.

\section{Experimental Results}
We first analyse the correlation of the ranks against the gold-standard order. Our novel metric achieves a perfect performance on both real and synthetic data. Further, three previous metrics, BLEU versus self-BLEU, LM score versus reverse LM score, and FED are evaluated accurately by adjusting the temperature. The results show that none of them are qualified as an unconditional text generation metric.

\subsection{Correlation Analysis on Real Scenario}
We rank the five LSTM-based generators and five GPT-2 based generators according to equation \ref{equ8} and the procedure described in Section 3, respectively. The gold-standard reference order is estimated, and given the architecture of the LM, a larger number of training data results in a better performance. Unlike the experiment on the syntactic data, we cannot obtain a real LM that generates real sentences. We did not compare these 10 generators together. Table \ref{Tb:performanceONreal} summarises the results.

\begin{table}[h]
  \caption{Distributional discrepancy of generators and classification accuracy on a real corpus. Acc. indicates the classification accuracy. For all of them, the lower the value, the better is the performance.}
   \label{Tb:performanceONreal}
   \begin{center}
   \begin{tabular}{|c|cc|cc|c|cc|cc|}
   \hline
   \multirow{2}{*}{Generator} &
    \multicolumn{2}{c|}{Valid} & \multicolumn{2}{c|}{Test} &
    \multirow{2}{*}{Generator} &
    \multicolumn{2}{c|}{Valid} & \multicolumn{2}{c|}{Test} \\
    \cline{2-5} \cline{7-10}
    & Acc. & DD & Acc. & DD
    &
    & Acc. & DD & Acc. & DD \\
   \hline
   $\text{LSTM}_{0.2}$ & 0.758 & 0.517 & 0.745 &0.490 & $\text{GPT-2}_{0.2}$ & 0.752 & 0.504 & 0.723 & 0.446 \\
  \hline
   $\text{LSTM}_{0.4}$ & 0.718 & 0.437 & 0.706 &0.412 & $\text{GPT-2}_{0.4}$ & 0.682 & 0.365 & 0.670 & 0.340 \\
  \hline
  $\text{LSTM}_{0.6}$ & 0.700 & 0.400 & 0.697 &0.393 & $\text{GPT-2}_{0.6}$ & 0.667 & 0.333 & 0.655 & 0.309 \\
  \hline
  $\text{LSTM}_{0.8}$ & 0.673 & 0.347 & 0.677 &0.353 & $\text{GPT-2}_{0.8}$ & 0.630 & 0.259 & 0.646 & 0.292 \\
  \hline
  $\text{LSTM}_{1.0}$ & 0.648 & 0.296 & 0.661 &0.322 & $\text{GPT-2}_{1.0}$ & 0.588 & 0.176 & 0.621 & 0.242 \\
  \hline

 \end{tabular}
 \end{center}
\end{table}

The rank achieved by the DD matches the gold-standard rank perfectly across the two architectures. For comparison of the previous metrics, Kendall's Tau co-efficiency is computed. Table \ref{Tb:TauReal} lists the results. Our novel metric still achieves the best performance. Although the LM score works considerably well for GPT-2, it fails to discriminate the generators when an LSTM architecture is adapted.

\begin{table}[h]
  \caption{Kendall's Tau rank correlation on a real corpus. 5LSTM and 5GPT-2 denote the correlations of five LSTM-based and five GPT-2 based generators, respectively.}
   \label{Tb:TauReal}
   \begin{center}
   \begin{tabular}{|c|cc|}
   \hline
   Metric  & 5LSTM & $\text{5GPT-2}$ \\
   \hline
   BLEU-5 & 0.0 & 0.6 \\
   \hline
   FED & 0.6 & 0.8 \\
   \hline
   LM score & -0.2  & \textbf{1.0} \\
   \hline
   DD & \textbf{1.0} & \textbf{1.0} \\
   \hline

 \end{tabular}
 \end{center}
\end{table}

Similar to synthetic data, we investigate the previous metrics in detail by adjusting the softmax temperature. All the evaluations shown in the above table were conducted under a condition of a temperature of 1.0.

\subsection{Correlation Analysis on Synthetic Data}
We evaluate all 10 generators based on their DD. The rank achieved by the DD matches the gold-standard order perfectly across the two models. All the $\tau$ values are listed in Table \ref{Tb:TauSyn}. 

\begin{table}[h]
  \caption{Kendall's Tau rank correlation on syntactic data. 10Gs denotes the Tau value when all 10 generators are evaluated together. 5LSTM and 5GPT-2 denote the correlations of five LSTM-based and five GPT-2-based generators, respectively.}
   \label{Tb:TauSyn}
   \begin{center}
   \begin{tabular}{|c|cc|c|}
   \hline
   Metric  & 5LSTM & 5GPT-2 & 10Gs\\
   \hline
   BLEU & 0.80 & 0.80 & 0.38\\
   \hline
   LM score & 0.40  & \textbf{1.0} & 0.82\\
   \hline
   FED & 0.84 & 0.84 & 0.92\\
   \hline
   DD & \textbf{1.0} & \textbf{1.0} & \textbf{1.0} \\
   \hline

 \end{tabular}
 \end{center}
\end{table}


Considering that the rank of the FED scores is similar to that of our own, we inspected its performance in detail. Table \ref{Tb:FED} lists all the scores. The discrimination is significantly small, and there is only a 0.006 difference between the best and worst scores. For the worst scores, several generators are assigned the same scores; however, their performances are different from each other.
\begin{table}[h]
  \caption{Ten generators are ranked according to FED score. The lower the score is, the better is the performance.}
  \scriptsize
   \label{Tb:FED}
   \begin{center}
   \begin{tabular}{|c|c|c|c|c|c|c|c|c|c|}
   \hline
    $\text{GPT-2}_{1.0}$ & $\text{GPT-2}_{0.8}$ & $\text{GPT-2}_{0.6}$ & $\text{GPT-2}_{0.4}$ &
    $\text{LSTM}_{1.0}$ & $\text{LSTM}_{0.8}$ & $\text{LSTM}_{0.6}$ & $\text{LSTM}_{0.4}$ & $\text{GPT-2}_{0.2}$ & $\text{LSTM}_{0.2}$\\
   \hline
    0.01 & 0.011 & 0.011 & 0.011 & 0.012 & 0.012 & 0.012 & 0.014 & 0.014 & 0.016\\
   \hline

 \end{tabular}
 \end{center}
\end{table}

To investigate the previous metrics in detail, we adjust the temperature of softmax. All evaluations shown in the above tables were conducted under the condition of a softmax temperature of 1.0.

\subsection{Detailed Analysis under Real Scenario}
\begin{figure*}[htbp]
\centering

\includegraphics[scale=0.58]{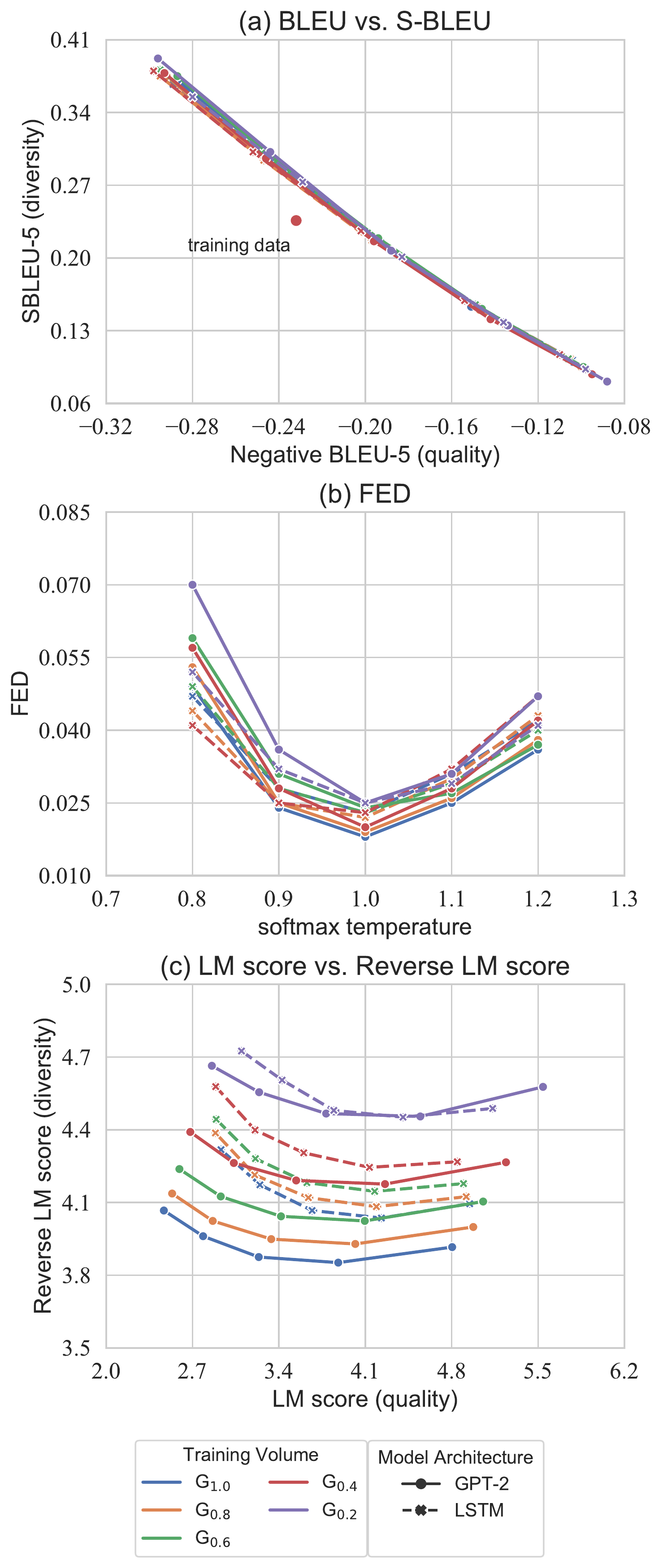}

\caption{Precise evaluation of generators using three previous metrics by adjusting the softmax temperature on a real corpus. For all of them, the lower the score, the better is the performance.}
\label{figEMNLP}
\end{figure*}

Following \citep{Caccia2019falling}, we evaluate these generators using three previous metrics by adjusting the softmax temperature. The temperature is set to 0.8, 0.9, 1.0, 1.1, and 1.2. Figure \ref{figEMNLP} illustrates the evaluation results. From these figures, we can see that both BLEU versus self-BLEU and FED fail to rank these generators.

Although the LM score versus reverse LM score ranked the five GPT-2 generators perfectly, the result was not as good for the five LSTM generators. This is inconvenient and inefficient because we must adjust the softmax temperature and draw the results in a two-dimensional metric. Further, this paired metric fails on synthetic data; this is illustrated in the next section.

In addition to the results of BLEU5 versus self-BLEU5, Table \ref{t1:emnlpresult} lists the detailed results of the other grams with a temperature of 1.0. As observed in this table, none of them can rank these generators as well as the DD.

\begin{table*}[t!]\footnotesize
\centering
\caption{More BLEU and self-BLEU results on EMNLP2017 WMT News. All results are run with five random initialisations, where $\varepsilon=0.001$. For self-BLEU, the lower the score, the better is the performance.
}
\begin{tabular}{|c|ccc|ccc|}
\hline
\multirow{2}{*}{Generator} &
\multicolumn{3}{c|}{BLEU} & \multicolumn{3}{c|}{Self-BLEU} \\
\cline{2-7}
& 2 & 3 & 4 & 2 & 3 & 4\\
\hline
$\text{LSTM}_{0.2}$ &
0.850$\pm 2\varepsilon$ & 0.587$\pm 4\varepsilon$ & 0.340$\pm 4\varepsilon$ &\textbf{ 0.863}$\pm 2\varepsilon$ & \textbf{0.616}$\pm 4\varepsilon$ & 0.\textbf{370}$\pm 6\varepsilon$\\
$\text{LSTM}_{0.4}$ &
\textbf{0.860}$\pm \varepsilon$ & 0.\textbf{607}$\pm \varepsilon$ & 0.\textbf{362}$\pm 2\varepsilon$ & 0.871$\pm \varepsilon$ & 0.633$\pm \varepsilon$ & 0.393$\pm 2\varepsilon$\\
$\text{LSTM}_{0.6}$ &
0.857$\pm \varepsilon$ & 0.601$\pm 2\varepsilon$ & 0.356$\pm 2\varepsilon$ & 0.869$\pm \varepsilon$ & 0.629$\pm 3\varepsilon$ & 0.389$\pm 3\varepsilon$\\
$\text{LSTM}_{0.8}$ &
\textbf{0.860}$\pm \varepsilon$ & 0.606$\pm \varepsilon$ & 0.360$\pm 2\varepsilon$ & 0.869$\pm \varepsilon$ & 0.631$\pm 3\varepsilon$ & 0.390$\pm 4\varepsilon$\\
$\text{LSTM}_{1.0}$ &
0.856$\pm \varepsilon$ & 0.598$\pm 2\varepsilon$ & 0.352$\pm 3\varepsilon$ & 0.867$\pm \varepsilon$ & 0.624$\pm 2\varepsilon$ & 0.383$\pm 3\varepsilon$\\
\hline
$\text{GPT-2}_{0.2}$ &
0.838$\pm 3\varepsilon$ & 0.577$\pm 5\varepsilon$ & 0.334$\pm 5\varepsilon$ & \textbf{0.850}$\pm 3\varepsilon$ & \textbf{0.600}$\pm 6\varepsilon$ & 0.\textbf{360}$\pm 7\varepsilon$\\
$\text{GPT-2}_{0.4}$ &
0.844$\pm 1\varepsilon$ & 0.590$\pm 3\varepsilon$ & 0.350$\pm 3\varepsilon$ & 0.856$\pm 2\varepsilon$ & 0.613$\pm 4\varepsilon$ & 0.379$\pm 4\varepsilon$\\
$\text{GPT-2}_{0.6}$ &
0.845$\pm 2\varepsilon$ & 0.591$\pm 3\varepsilon$ & 0.351$\pm 2\varepsilon$ & 0.858$\pm 2\varepsilon$ & 0.618$\pm 4\varepsilon$ & 0.383$\pm 5\varepsilon$\\
$\text{GPT-2}_{0.8}$ &
\textbf{0.850}$\pm 1\varepsilon$ &\textbf{ 0.601}$\pm 2\varepsilon$ & \textbf{0.361}$\pm 2\varepsilon$ & 0.862$\pm 1\varepsilon$ & 0.627$\pm 5\varepsilon$ & 0.392$\pm 2\varepsilon$\\
$\text{GPT-2}_{1.0}$ &
0.849$\pm 3\varepsilon$ & 0.598$\pm 5\varepsilon$ & 0.358$\pm 5\varepsilon$ & 0.862$\pm 2\varepsilon$ & 0.624$\pm 4\varepsilon$ & 0.390$\pm 5\varepsilon$\\
\hline

\end{tabular}

\label{t1:emnlpresult}
\end{table*}

\subsection{Detailed Analysis on Synthetic Data}
\begin{figure*}[htbp]
\centering

\includegraphics[scale=0.58]{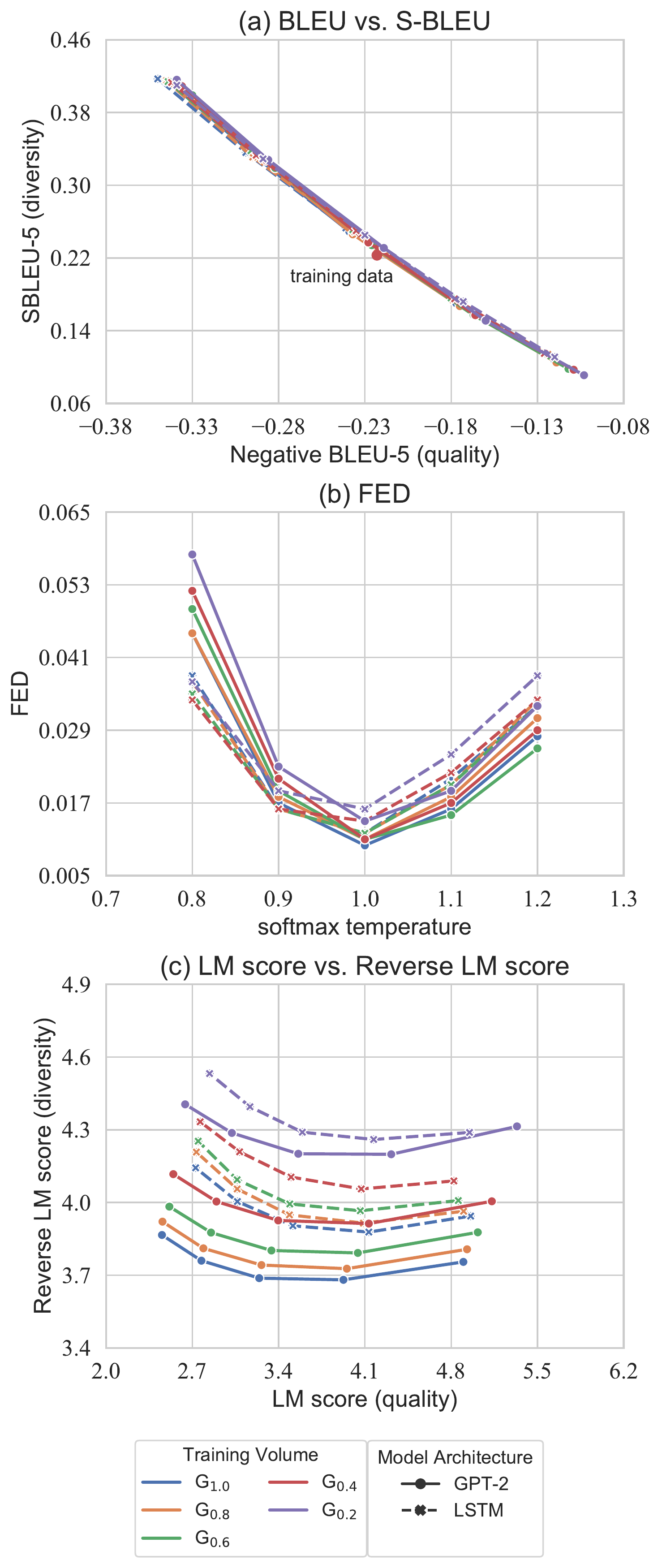}

\caption{Precise evaluation of generators using three previous metrics by adjusting the softmax temperature on synthetic data. For all of them, the lower the score, the better is the performance.}
\label{figSynthetic}
\end{figure*}

The same settings of the softmax temperature as in the real data are adapted. Unlike for the real data, we compare all 10 generators together. Figure \ref{figSynthetic} shows that BLEU versus self-BLEU and FED fail to discriminate these generators even when they have the same architecture. The ranks that are achieved by LM score versus reverse LM score on the same architecture generators are consistent with the gold-standard. However, they cannot rank the 10 generators well because a finer calibration is required.

Surprisingly, our single metric ranks all 10 generators considerably well, as shown in Table \ref{Tb:TauSyn}. It is not necessary to adjust the generators' softmax temperature. This shows that DD is powerful and efficient.

\section{Conclusion and Future work}
We present a novel metric, distributional discrepancy, to measure the discrepancy between real and generated texts unconditionally. A neural network classifier is trained to classify the true and generated texts. We exploit the classification accuracy to obtain this discrepancy.

Compared with the existing metrics, this single metric can clearly distinguish the different generative models and evaluate them in terms of both quality and diversity of the sample. Numerous experiments show that distributional discrepancy perfectly ranks two architecture models on both synthetic data and a real corpus. However, the existing metrics either cannot evaluate these generative models or are inefficient.

In the future, a stronger classifier such as LSTM and transformer will be investigated to verify the robustness of this metric. Further, a larger-scale corpus such as wiki-103 \citep{Merity2016Wiki103} will be tested.

\section*{Acknowledgment}
This research was supported by the National Key R\&D Program of China (No. 2017YFB1401401), and Nature Science Foundation of China (No. 81373056), and partly by the Key program for International S\&T Cooperation of Sichuan Province (No. 2019YFH0097). We thank the anonymous reviewers for their helpful comments and suggestions.


\bibliography{mybibfile}

\end{document}